\documentclass[journal,letters]{IEEEtran}

\usepackage[final]{graphicx}
\usepackage[reqno]{amsmath}
\usepackage{amssymb}
\usepackage{amsthm}
\usepackage{subfig}
\usepackage{algorithm}
\usepackage{algorithmicx}
\usepackage{algpseudocode}
\usepackage{setspace}
\usepackage[T1]{fontenc}
\usepackage{color,soul}
\usepackage{multirow}
\usepackage{array}
\usepackage{multicol}
\usepackage{graphicx}
\usepackage[algo2e,linesnumbered,ruled,vlined]{algorithm2e}
\usepackage{float}
\usepackage{url}
\usepackage{booktabs}
\usepackage{subfig}
\usepackage{caption}

\begin{document}

\title{Efficient Training of Deep Classifiers for Wireless Source Identification using Test SNR Estimates}

\author{\large{Xingchen Wang, {\em Student Member, IEEE}}, \large{Shengtai Ju, {\em Student Member, IEEE}}, \large{Xiwen Zhang, {\em Student Member, IEEE}}, \large{Sharan Ramjee, {\em Student Member, IEEE}}, and \large{Aly El Gamal, {\em Senior Member, IEEE} }
\thanks{X. Wang, S. Ju, X. Zhang, S. Ramjee, and A. El Gamal are with the Department of Electrical and Computer Engineering, Purdue University, West Lafayette, IN, USA. Email: \{wang2930,  ju10, zhan2977, sramjee, elgamala\}@purdue.edu.}}

\maketitle

\begin{abstract}
We study efficient deep learning training algorithms that process received wireless signals, if a test Signal to Noise Ratio (SNR) estimate is available. We focus on two tasks that facilitate source identification: 1- Identifying the modulation type, 2- Identifying the wireless technology and channel in the 2.4 GHz ISM band. For benchmarking, we rely on recent literature on testing deep learning algorithms against two well-known datasets. We first demonstrate that using training data corresponding only to the test SNR value leads to dramatic reductions in training time while incurring a small loss in average test accuracy, as it improves the accuracy for low SNR values. Further, we show that an erroneous test SNR estimate with a small positive offset is better for training than another having the same error magnitude with a negative offset. Secondly, we introduce a greedy training SNR Boosting algorithm that leads to uniform improvement in accuracy across all tested SNR values, while using a small subset of training SNR values at each test SNR. Finally, we demonstrate the potential of bootstrap aggregating (Bagging) based on training SNR values to improve generalization at low test SNR values with scarcity of training data. %the potential of deep learning regularization and optimization algorithms, that are based on training set selection, like Bagging and curriculum learning for accuracy improvement, specially at low test SNR values with scarcity of training data.  
\end{abstract}

%\begin{IEEEkeywords}
%Deep Learning for Wireless, Modulation Classification, Channel Identification, SNR Boosting, SNR Bagging.
%\end{IEEEkeywords}

\IEEEpeerreviewmaketitle

\begin{table*}
\centering
\caption{Architectures for Modulation Classification (MC) and Channel Identification (CI). The convolutional layers column indicates the number of kernels and their size for each layer. The dense layers column indicates the input and output size of each layer. The LSTM column indicates the number of recurrent cells. The last column indicates the number of residual stacks.} 

    \begin{tabular}{|c|c|c|c|c|c|c|}   
    \hline
    Architecture & Activation Functions & Convolutional Layers & Dense Layers & LSTM & R-Stacks\\    
    \hline
    CNN (MC)  & ReLU, Softmax & 256 $3*1$, 80 $3*2$ & $10560*256$, $256*10$ &  &   \\
    \hline
    \multirow{2}{*}{ResNet (MC)} & \multirow{2}{*}{ReLU, SeLU, Softmax} &  & $128*128$, $128*128$, &  & \multirow{2}{*}{5}   \\
     &  &  &  $128*10$ &  &    \\
    \hline
    \multirow{2}{*}{CLDNN (MC)}  & \multirow{2}{*}{ReLU, Softmax} & 256 $3*1$, 256 $3*2$, & \multirow{2}{*}{$50*256$, $256*10$}  & \multirow{2}{*}{50} &  \\
      &  & 80 $3*1$, 80 $3*1$ &   & &  \\
    \hline
   CNN-1 (CI) & ReLU, Softmax & 256 $3*1$, 256 $3*2$ & $31744*1024$, $1024*15$ &  &   \\
    \hline
    CNN-2 (CI) & ReLU, Softmax & 256 $3*1$, 256 $3*1$ & $32768*1024$, $1024*15$ &  &   \\
    \hline
    \multirow{2}{*}{ResNet (CI)} & \multirow{2}{*}{ReLU, Softmax} &  & $128*128$, $128*128$, &  & \multirow{2}{*}{5}   \\
    &  &  & $128*15$ &  &    \\
    \hline
    \multirow{2}{*}{CLDNN (CI)}  & \multirow{2}{*}{ReLU, Softmax} & \multirow{2}{*}{256 $3*1$, 256 $3*2$} &\multirow{2}{*}{$31744*1024$ (before LSTM),} & \multirow{2}{*}{256} &\\&&&&&\\& & & $512*256$, $256*15$  & (2-dim)  &\\
    \hline
    \end{tabular}

\label{table: neural network architecture for modulation}
\end{table*}
    
%\begin{table*}
%    \centering
%    \caption{The considered deep neural network architectures for channel identification.}
%    \begin{tabular}{|c|c|c|c|c|c|c|}   
%    \hline
%    Architecture & Activation Functions & Convolutional Layers & Dense Layers & LSTM & R-Stacks\\    
%    \hline
%    CNN-1 & ReLU, Softmax & 256 $3*1$, 256 $3*2$ & $31744*1024$, $1024*15$ &  &   \\
%    \hline
%    CNN-2 & ReLU, Softmax & 256 $3*1$, 256 $3*1$ & $32768*1024$, $1024*15$ &  &   \\
%    \hline
%    \multirow{2}{*}{ResNet} & \multirow{2}{*}{ReLU, Softmax} &  & $128*128$, $128*128$, &  & \multirow{2}{*}{5}   \\
%    &  &  & $128*15$ &  &    \\
%    \hline
%    \multirow{2}{*}{CLDNN}  & \multirow{2}{*}{ReLU, Softmax} & \multirow{2}{*}{256 $3*1$, 256 $3*2$} &\multirow{2}{*}{$31744*1024$ (before LSTM),} & \multirow{2}{*}{256} &\\&&&&&\\& & & $512*256$, $256*15$  & (2-dim)  &\\
%    \hline
%    \end{tabular}
%\label{table: neural network architecture for identification}
%\end{table*}

\begin{table*}[ht]
\begin{center}
\caption{Average accuracy and training times with best performing architectures for each task.}
\begin{tabular}{ |c|c|c|c|c|c|} 
 \hline
 Architecture & Training data type & Accuracy (\%) & Time per epoch (s) & Number of epochs & Training time (s) \\ 
 \hline
 \multirow{2}{*}{ResNet (MC)} & Single SNR & 60.46 & 1.00 & 49.65 & 49.65 \\
 \cline{2-6}
 & All SNR & 63.00 & 27.50 & 48.00 & 1320.00 \\ 
 \hline
 \multirow{2}{*}{CNN (CI)} & Single SNR & 90.19 & 0.98 & 34.52 & 33.84 \\
 \cline{2-6}
 & All SNR & 89.69 & 22.02 & 12.00 & 264.24 \\
 \hline
\end{tabular}

\label{table:single_mod_table}
\end{center}
\end{table*}

\vspace{-4 mm}
\section{Introduction}
%Deep Learning for Wireless and why Source Identification is important
\IEEEPARstart{D}{eep} learning can potentially become an essential part in the design of next generation wireless networks, because of the difficulty of modeling their environments as well as the small time scale of wireless communications that allows for rapidly collecting large datasets. More specifically, deep learning algorithms will be strong candidates for autonomous communication systems that require little computational and control overhead, and form an intelligent understanding of the spectrum, which starts with the basic task of identifying the source(s) of transmission for a given received wireless signal. 

%Modulation Classification dataset and literature
Recognizing the modulation type of a received signal is important for interference source identification, and for reducing control overhead by enabling frequent modulation and coding scheme adaptation to the changing environment. In \cite{datagen} and \cite{conv}, a synthetic dataset based on the GNU Radio software package was introduced to initiate the investigation of deep neural network algorithms that are suitable for recognizing one out of 10 modulation types, including both analog and digital types. In \cite{mod1}, multiple architectures were presented that deliver state of the art accuracy results for this synthetic dataset. In \cite{mod2}, fast training algorithms for these architectures were studied, and promising results were presented. %using candidate algorithms that rely on training with only a subset of the samples available in each input vector (subsampling) as well as a subset of vectors that correspond to only a single value for the signal to noise ratio (SNR). 

%Interference Identification dataset and literature
In \cite{schmidt2017wireless}, the problem of channel identification in the 2.4 GHz ISM band was considered through a synthetic dataset that the authors introduced. The dataset contains received signals corresponding to 15 different channels of WiFi, Bluetooth, and ZigBee. Efficient training algorithms for this problem were investigated in \cite{ident1}. % through training with only a subset of frequency bands, a subset of input vector samples, and a subset of SNR values. 
Also, in \cite{access}, an experimental framework for machine-learning-based channel identification using Berkeley TelosB sensorboards was studied. 

%Proposed approach and summary of results and added value to state of the art
In this work, we demonstrate the feasibility of effective and fast training of deep classifiers for wireless source identification tasks, in presence of a good test SNR estimate. We use the aforementioned datasets to validate the proposed methods and obtained insights. The contributions of this work can be summarized as follows:
1) We improve the preliminary results on training SNR selection of \cite{mod2} and \cite{ident1}, obtained through training with only the test SNR value, which constitutes only 5\% of available training data. We show that  the training time can be reduced by more than 30x, while incurring minimal loss in average accuracy. Further, the accuracy at low SNR values noticeably increases when using SNR selection. It is important to note that in practice, the SNR value can frequently undergo small changes, and hence, it is important to understand design guidelines for deep learning training algorithms in presence of erroneous test SNR estimates, which motivates our next contribution. 2) We study the sensitivity of obtained results to test SNR estimation errors.  We show that if we can only estimate a test SNR range, it is better to train with optimistic estimates. It is worth mentioning here that the impact of other estimation imperfections like erroneous sample rate and center frequency estimates, on the outcome of deep modulation classifiers have been investigated in \cite{milcom}. 3) We present a training \emph{SNR Boosting} algorithm that identifies a small set of training SNR values that lead to the best accuracy for every given test SNR. Not only does it lead to faster training, but the identified training set through SNR Boosting is shown to consistently deliver superior performance to training with all available data. 4) We investigate the potential of training SNR Bootstrap Aggregating (Bagging) by demonstrating significant performance improvements through reducing the generalization error in presence of low SNR and scarce training data.
We note that the effect of the SNR on the testing accuracy of supervised classifiers has been investigated in \cite{milcom} and \cite{access} for the considered tasks. Further, in these works, it is shown that the number of samples per packet and quantization error are important factors that contribute to the needed training time for a given accuracy guarantee. However, the distinguishing aspect of this work is investigating how knowledge about the test SNR can be useful in choosing the training set such that both the training time and classification accuracy are optimized.

%curriculum learning and Bagging (see e.g., \cite[Chapters 7 and 8]{dl-book}).

\begin{figure*}[ht]
\begin{multicols}{2}
    \includegraphics[width=\linewidth,height=4.5cm]{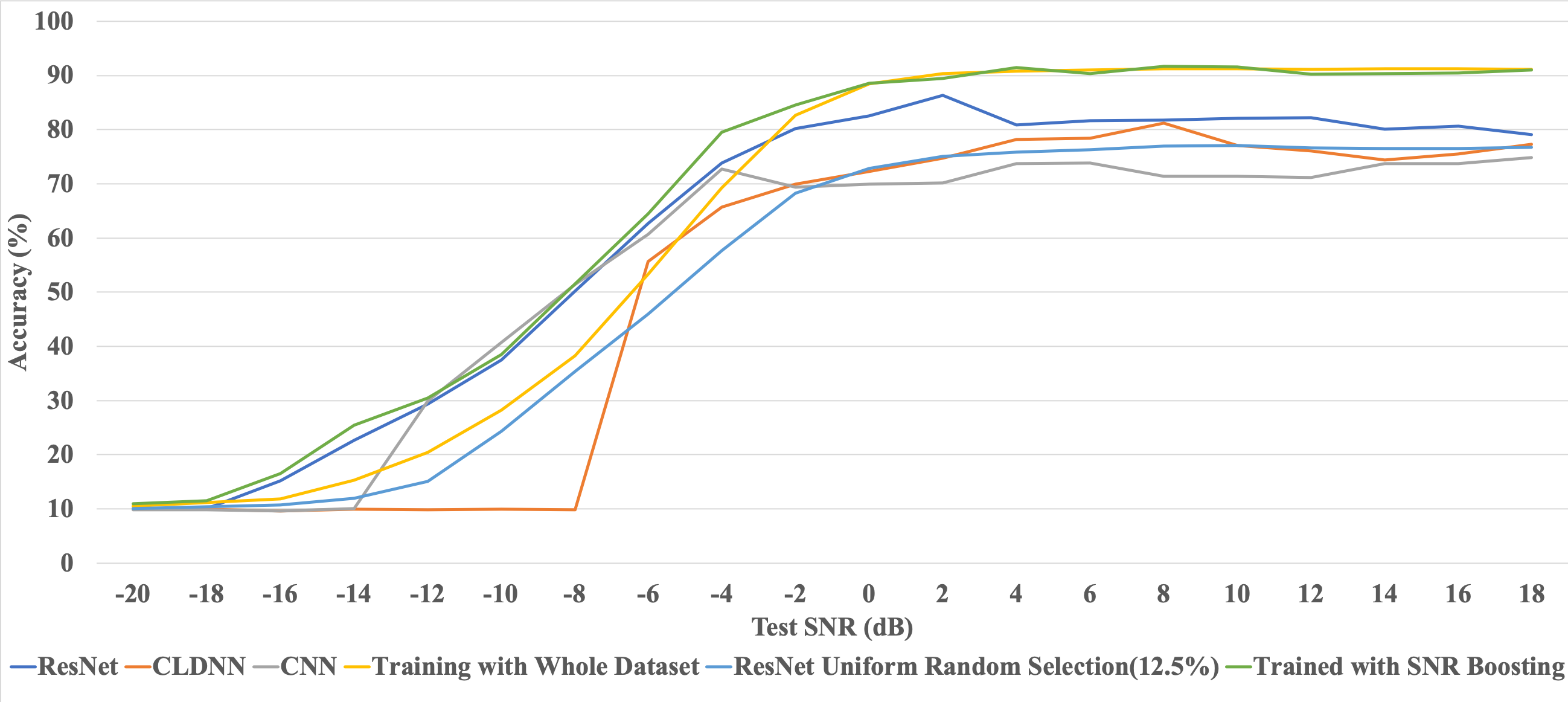}\par 
    \caption{Single SNR Selection with Considered Architectures and SNR boosting (ResNet) for Modulation Classification (ResNet is used for "Training with Whole Dataset").}
    \label{fig_single:single_mod_multi} 
    \includegraphics[width=\linewidth,height=4.5cm]{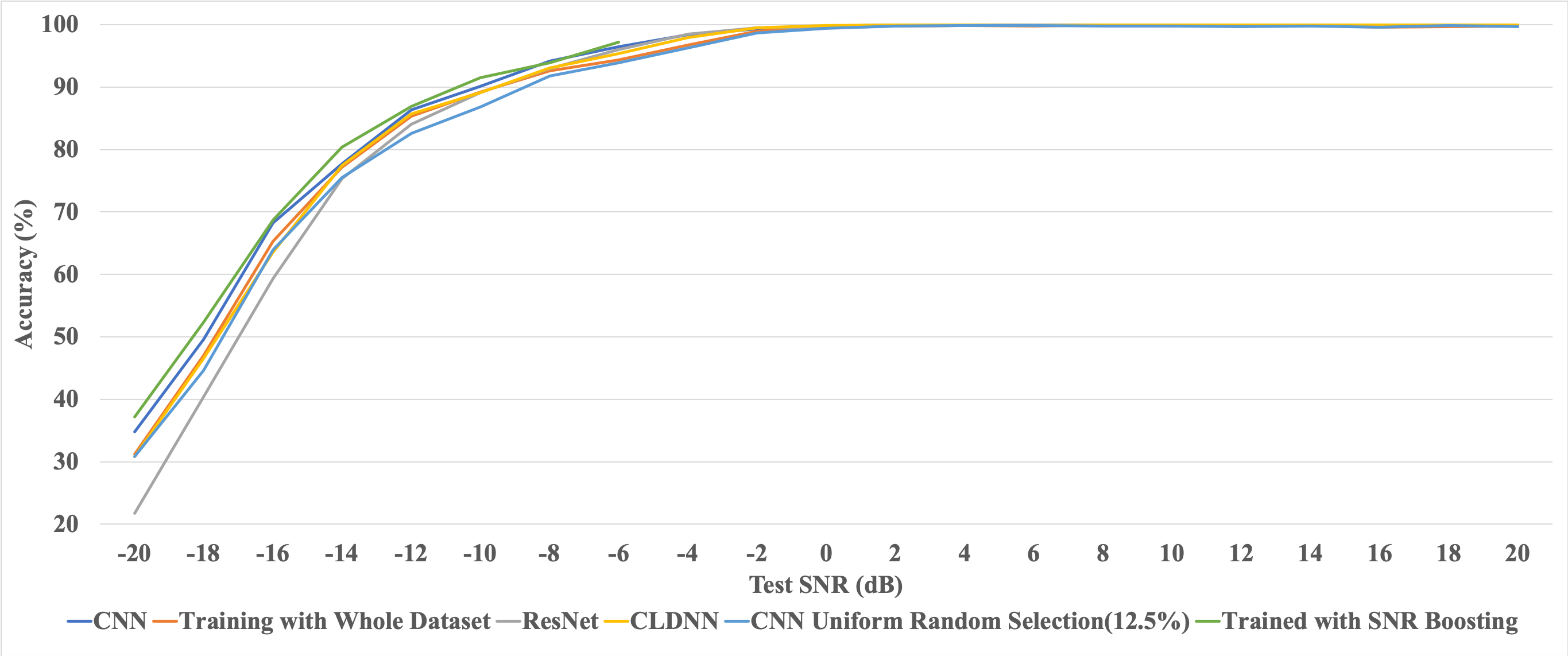}\par
     \caption{Single SNR Selection with Considered Architectures and SNR boosting (CNN) for Channel Identification (CNN is used for "Training with Whole Dataset").}
    \label{fig_single:single_inter_multi}
    \end{multicols}
\begin{multicols}{2}
    \includegraphics[width=\linewidth,height=4.5cm]{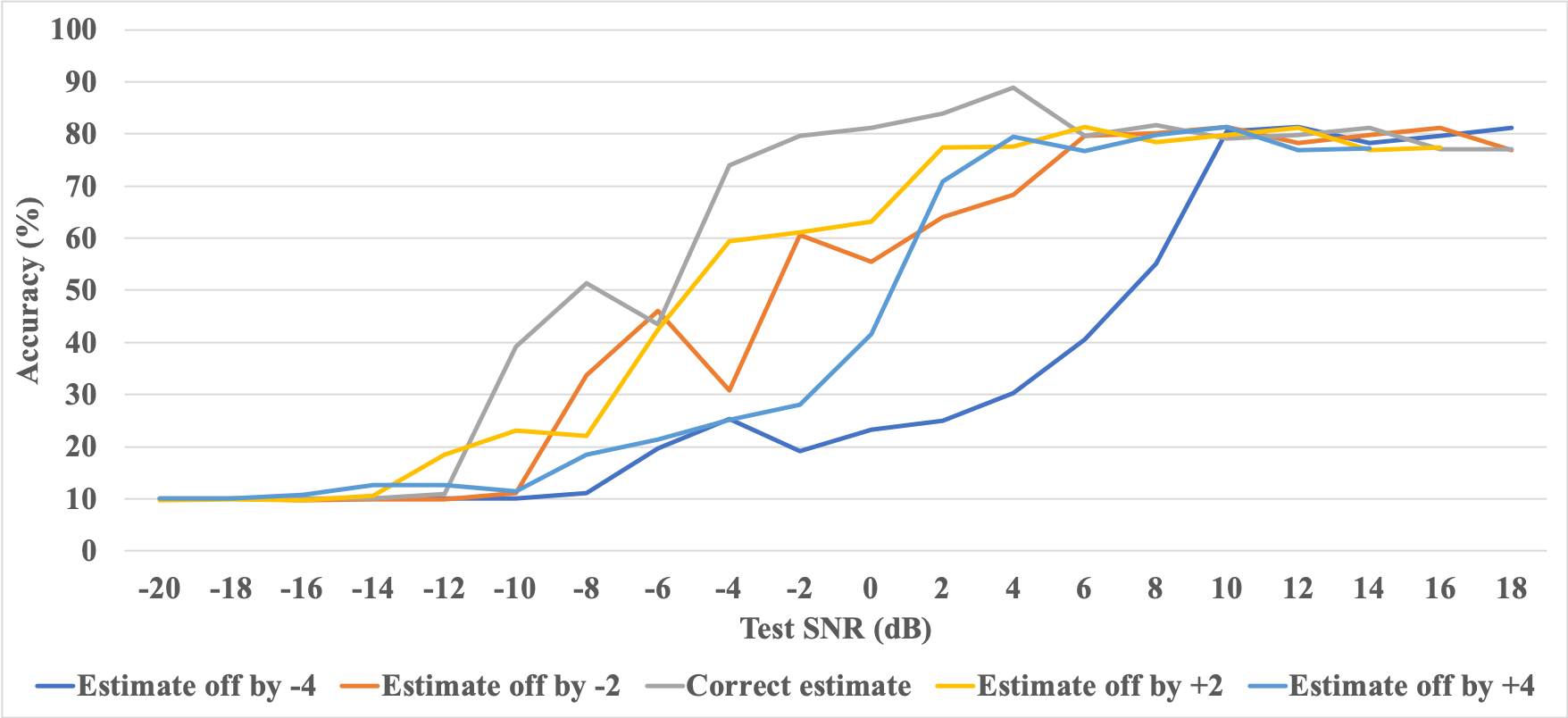}\par 
    \caption{ResNet SNR Sensitivity for Modulation Classification.}
    \label{fig_single:sensitivity_mod}
    \includegraphics[width=\linewidth,height=4.5cm]{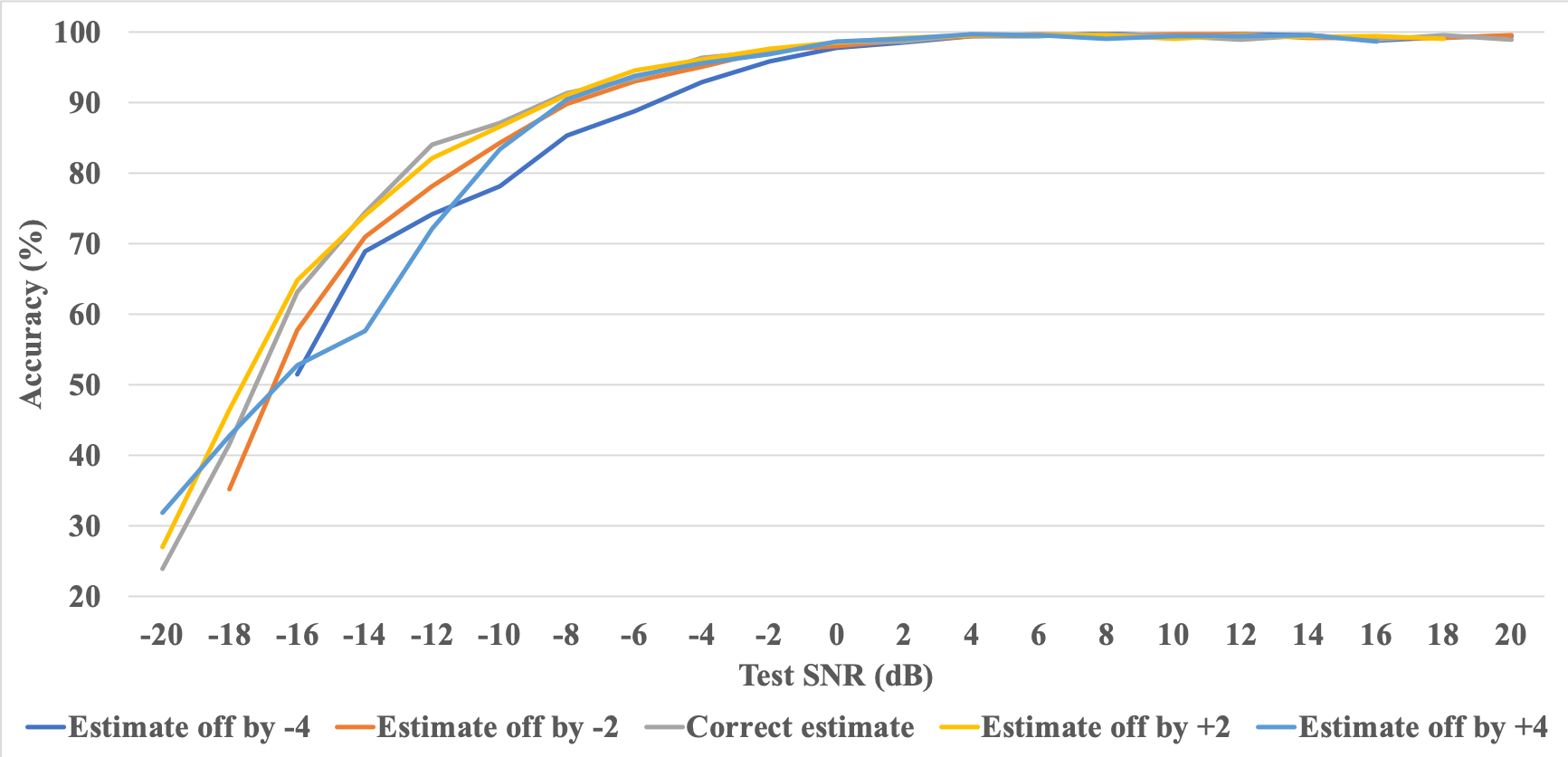}\par 
    \caption{SNR Sensitivity with CNN for Channel Identification.}
    \label{fig_single:sensitivity_inter}
\end{multicols}
%\begin{multicols}{2}
%    \includegraphics[width=\linewidth,height=4.5cm]{new_dataset.png}\par
%    \caption{ResNet SNR Sensitivity using RadioML2018.01A.}
%    \label{fig_single:new_dataset}
%    \includegraphics[width=\linewidth,height=4.5cm]{new_greedy_set.png}\par 
%    \caption{SNR Set Selected by Boosting Algorithm.}
%    \label{fig_single:greedy_set}
%\end{multicols}
\label{fig_single} 
\end{figure*}
\section{Problem Setup}\label{sec:setup}

%\subsection{Problem Description and Considered Training Algorithms}
\textbf{Problem Description:} Given a received signal, our goal is to maintain a high classification accuracy while driving the total training time as low as possible for computational efficiency\footnote{Code available at https://codeocean.com/capsule/7188167/tree/v1 (modulation classification) and https://codeocean.com/capsule/0a545fca-d0c3-410b-bc37-f253e585ac39/tree (channel identification)}. We consider two classification tasks: 1- Identifying one out of 10 modulation types for the RadioML2016.10b dataset of \cite{datagen}. 2- Identifying one out of 15 channels for the channel identification dataset of \cite{schmidt2017wireless}. Both datasets are synthetic and based on simulating random channel and hardware imperfections, that are difficult to model in closed form, and the descriptions of the random generators of these imperfections are available in \cite{datagen} and \cite{schmidt2017wireless}. Also, the noise model for both datasets is an Additive White Gaussian Noise (AWGN) model. 

\textbf{Considered Training Algorithms:} We consider the following three training scenarios: (a) Training with all available data. (b) Training only with data of the same SNR value as a test SNR estimate. (c) Training with a selected subset of available data corresponding to specific SNR values, which are determind through an SNR Boosting algorithm.
%We want to be able to predict the types of modulation of incoming signals to our system. We would like to maintain a high prediction accuracy while keeping the total training time and cost as low as possible for computational efficiency. Therefore, we consider the following three training scenarios: (a) training with all available data, (b) training only with data of the same SNR value as the target test SNR, (c) training with a selected subset of available data. Training with all available data is simply training a neural network with all of the data that is available to us, which contains all SNR values. Training with a single SNR requires the target test SNR to be given or accurately estimated. After the target SNR is known, we can pre-process the training data and only train the neural network with training data that has the same SNR as the target signal. We suspect that training only with data that has a single SNR value is not enough to get the optimal prediction accuracy. Therefore, we propose a novel algorithm called the SNR Boosting algorithm which selects a subset of training data given the target test SNR. By selecting a subset of the training data, we can potentially achieve high accuracy while keeping the training time low to save computation cost.
%\subsection{Datasets}\label{sec:dataset}

\textbf{Modulation Classification:} Ten widely used modulation types are chosen; eight digital and two analog. %These consist of BPSK, QPSK, 8PSK, QAM16, QAM64, BFSK, CPFSK, and PAM4 for digital modulations, and WB-FM, and AM-DSB for analog modulations.
The modulation types are listed in the confusion matrix of Figure \ref{fig_conf}. The dataset consists of 160,000 sample vectors; each consisting of 128 2-dimensional (real and imaginary) samples, taken every $1$  $\mu s$ from a baseband received signal, with a modulation rate of 8 samples per symbol. The sampling rate for this dataset is typically around 6 times the Nyquist rate as illustrated in \cite{mod2}.

\textbf{Channel Identification:} We have 225,225 sample vectors for 15 classes.
%\footnote{The dataset is available at \url{https://crawdad.org/owl/interference/20180925/}}
There are 10 1 MHz wide Bluetooth channels with center frequencies in the range 2422-2431 MHz, spaced every 1 MHz. Also, there are three 20 MHz wide WiFi channels with center frequencies of 2422, 2427, and 2432 MHz. Finally, there are two 2 MHz wide Zigbee channels with center frequencies 2425 and 2430 MHz. For WiFi, the Physical Layer Mode is varied between CCK, PBCC, and OFDM.
For Bluetooth, the Transport Mode is varied between ACL, SCO, and eSCO.
For Zigbee, the ACK-frame is used.
Each sample vector consists of 128 I/Q samples, corresponding to $12.8$  $\mu s$, and the captured band is 2421.5-2431.5 MHz. 
The I/Q samples of each sample vector are also transformed into the frequency domain by using the Fast Fourier Transform, as we have found this transformation to deliver better results than using time-domain data. For example, at 0 dB, it results in an accuracy increase with single SNR training from $94.78\%$ to $99.72\%$. We believe that this is because it facilitates identifying the channel through the occupied frequency range.
%\begin{table}\label{tab:classes}
%    \centering
%    \caption{The considered 15 classes of channels.}
%    \begin{tabular}{|c|c|c|}   
%    \hline
%    Technology & Center Frequency & Channel Width \\    
%    \hline
%    Bluetooth  & 2422 MHz         & 1 MHz  \\
%    \hline
%    Bluetooth  & 2423 MHz         & 1 MHz  \\
%    \hline
%    Bluetooth  & 2424 MHz         & 1 MHz  \\
%    \hline
%    Bluetooth  & 2425 MHz         & 1 MHz  \\
%    \hline
%    Bluetooth  & 2426 MHz         & 1 MHz  \\
%    \hline
%    Bluetooth  & 2427 MHz         & 1 MHz  \\
%    \hline
%    Bluetooth  & 2428 MHz         & 1 MHz  \\
%    \hline
%    Bluetooth  & 2429 MHz         & 1 MHz  \\
%    \hline
 %   Bluetooth  & 2430 MHz         & 1 MHz  \\
%    \hline
%    Bluetooth  & 2431 MHz         & 1 MHz  \\
%    \hline
%    WiFi       & 2422 MHz         & 20 MHz \\
%    \hline
%    WiFi       & 2427 MHz         & 20 MHz \\
%    \hline
%    WiFi       & 2432 MHz         & 20 MHz \\
%    \hline
%    Zigbee     & 2425 MHz         & 2 MHz  \\
%    \hline
%    Zigbee     & 2430 MHz         & 2 MHz  \\
%    \hline
%    \end{tabular}
%\end{table}
Both datasets are split equally among all classes and SNR values from -20 dB to 18 dB for modulation classification and to 20 dB for channel identification, and a step size of 2 dB.

\section{Deep Neural Network Algorithms}

We consider three different architecture types: A Convolutional Neural Network (CNN), Residual Network (ResNet), and a Convolutional Long Short-term Deep Neural Network (CLDNN). The key motivations for our choice of architectures is to exploit the powerful advantage of parameter sharing available when using convolutional kernels and the ability of Long Short-Term Memory (LSTM) cells to capture temporal correlations in long sequences (see \cite[Chapters 9-10]{dl-book}). Further, ResNets can enhance generalization by allowing for stable optimization of deeper networks through shortcut connections. We used a GPU server with a Tesla P100 GPU and 16 GB of memory, and the optimization algorithms specified in \cite{mod2, ident1}.

\textbf{Modulation Classification:} The CNN consists of two convolutional layers by first having 256 $3 \times 1$ kernels, and then 80 $3 \times 2$ kernels. The two dense layers have sizes of $1024 \times 256$ and $256 \times 10$, respectively. Note that we chose the smallest kernels that led to good performance to reduce training computational cost, and that the choice of more kernels in earlier layers emanates from the assumption that higher-level patterns - found in deeper layers - are fewer than lower-level ones. Activation functions used in the CNN are ReLU for hidden layers and Softmax for the output layer (see \cite[Chapter $6$]{dl-book} for rationale). The ResNet architecture used for single SNR selection contains three residual stacks. A residual stack consists of the following: a 1-D convolutional layer with kernel size 1 and linear activation function followed by a batch normalization layer; after batch normalization, two residual units are added, and finally, a max pooling layer is added. Note that linear activations in deeper networks are specially beneficial as they guide the parameter optimization towards a lower dimensional subspace (see \cite[Chapter 6]{dl-book}). The residual unit contains two convolutional layers with kernel size 5, each followed by a batch normalization layer. The first convolutional layer in the residual unit uses ReLU activation while the second uses a linear activation function. A shortcut unit is added for the residual unit connecting the beginning and the end. The ResNet architecture used in our boosting algorithm has five residual stacks with the exact same setup just mentioned. The dimension of the first dense layer for our 3-stack ResNet is $512$ and the dimension of the first dense layer for our 5-stack ResNet is $128$. It is important to note here that we chose a deeper and narrower ResNet for boosting, as this can greatly enhance generalization when training is aided by an external process that provides side information about the right hyperparameter space to restrict the search to (see \cite[Chapters 6 and 7]{dl-book} for more details). The ResNet dense layers have Scaled Exponential Linear Unit (SeLU) activation. ResNet stack architectures are inspired by popular choices that proved to be successful for multiple applications (see e.g., \cite{resnet18}). For the CLDNN, the LSTM layer follows all convolutional layers, and precedes all dense layers (see \cite[Chapter $10$]{dl-book}). The position of this layer is chosen to create a \emph{summary} of features learned from applying convolution. %while capturing long term temporal correlations.

%\subsection{Channel Identification}
\textbf{Channel Identification:} For the CNN used for single SNR selection, a dropout layer is added after the second convolutional layer. The output is then reshaped and fed to a fully connected layer followed by ReLU and another dropout layer. For SNR boosting, batch normalization is applied after each convolutional layer, and the dropout layer after the second convolution layer is removed. Note that the extra dropout layers for SNR selection are needed for improving generalization due to the smaller training set. Also, reducing regularization by removing a dropout layer for boosting is intuitive as the boosting algorithm can guide the training procedure towards a smaller space of possible parameter settings. The CNN architectures used for single SNR selection and SNR Boosting are labeled CNN-1 and CNN-2 in Table \ref{table: neural network architecture for modulation}, respectively. The ResNet used for channel identification has five residual stacks. The residual stacks are similar to those used for modulation classification, but both convolutional layers in a residual unit have ReLU activations. For the CLDNN, one dense layer succeeds convolutional layers and precedes the LSTM layer, and the other two dense layers follow the LSTM layer. %The positioning of the LSTM layer follows a similar logic to that explained above. 
%Each residual stack has the following: a convolutional layer, a batch normalization layer, two residual units, and a max pooling layer. Each residual unit has two convolutional layers with kernel size 5 each followed by a batch normalization layer. Both convolutional layers in the residual unit use ReLu as activation function. A shortcut is added which sums the input with the output directly. Details of all the architectures used can be found in Table \ref{table: neural network architecture for identification}. 

%\subsection{Programming Environment and Hyperparameters}\label{sec:hardware}
%\textbf{Programming Environment and Hyperparameters:} We used a GPU server with a Tesla P100 GPU and 16 GB of memory. %For modulation classification, we used Keras version 2.2.4 with TensorFlow backend, and for channel identification, we used PyTorch version 1.3.1.  
%For modulation classification, we used the same hyperparameters and optimization algorithms as in \cite{mod2}. For channel identification, we used a dropout rate of 0.6 for single SNR selection, and 0.8088 for greedy SNR boosting. All other parameters are kept the same as \cite{ident1}. 

\section{Results}
%\subsection{Single SNR selection}\label{sec:single}
%\subsubsection{Modulation Classification}
%Main Result
\textbf{Single SNR Selection for Modulation Classification:} When training with a perfect test SNR estimate, we observe from Figure \ref{fig_single:single_mod_multi} that the accuracy is typically higher than using the whole training set at low SNR values, and lower at high SNR values. The average accuracy suffers from a slight drop as illustrated in Table \ref{table:single_mod_table}. The intuition behind the low SNR improved performance is that when training with the whole dataset, the network focuses on patterns irrelevant to the noisy regime corresponding to the test SNR. %We note that the superior relative performance of SNR selection at low SNR values can also be observed in Figure~\ref{fig_single:new_dataset} for the larger dataset.  
%why lower SNR is better - importance of estimate
As we are using 5\% of the training dataset, the training time is reduced by 25-35x (roughly on average 27x for ResNet and CNN, and 35x for CLDNN). In the table, an \emph{epoch} refers to a complete pass over the training set while feeding the example batches (see \cite[Chapter 8]{dl-book}). The training time reduction is typically larger than the reduction in size of the training set, and hence training all \emph{single SNR classifiers} would require less time than training a single classifier with the whole dataset. Further, these single SNR classifiers can be trained independently in parallel. This can be beneficial in practical scenarios where online training and calibration are frequently needed using fresh data. We also observe from Figure \ref{fig_single:single_mod_multi} that when using a larger portion of 12.5\% of the training dataset, but uniformly distributed across all SNR values, the test accuracy is uniformly (at all test SNR values) lower than the single SNR selection strategy.
%training time reduction 25-35x
%Comparison with uniform

We restrict our attention to the ResNet for the remaining experiments since it was found to deliver the best performance. In order to determine the training SNR selection strategy when the test SNR estimate can only specify a range of values, we tested the impact of training with an SNR that is lower/upper than the true value by 2 and 4 dB. As shown in Figure~\ref{fig_single:sensitivity_mod}, the optimistic estimate is almost always better to train with. 
%Estimation Errors
%training with cleaner data is better

%To maintain high prediction accuracy while minimizing training time, we pre-process the training data to split the data into subsets with only single SNR values, ranging from $-20$ dB to $20$ dB. We then train different neural network models with single-SNR training data. From Table \ref{table:single_mod_table}, we can see that single SNR selection leads to drastic decrease in training time across all architectures while maintaining high classification accuracy. We choose ResNet to study in further sections because ResNet outperforms all other architectures. Single SNR selection reduces the size of training dataset to its $\frac{1}{20}$. Compared to uniform subsampling which reduces the size of training dataset to its $\frac{1}{8}$, Figure \ref{fig_single:single_mod_multi} shows that single SNR selection maintains higher accuracy while requiring less training data which leads to less training time. We also consider the case where we cannot estimate the target SNR accurately. By comparing negative SNR offsets in SNR estimation and positive SNR offsets in SNR estimation, Figure \ref{fig_single:sensitivity_mod} shows that estimates off by some positive SNR are significantly better than those with negative SNR offsets.
%\subsubsection{Channel Identification}
\textbf{Single SNR Selection for Channel Identification:} Identical observations to the modulation classification task hold, with the following exceptions: 1- Starting from test SNR of 0 dB, we obtain almost perfect classification accuracy, and hence, the stated observations are evident only at lower test SNR values. We believe that this difference is due to the easier classification task, and not a result of the higher sampling rate, because while the sampling rate for channel identification is the Nyquist rate for WiFi signals that occupy the whole considered 10 MHz bandwidth, the sampling rate for modulation classification is above the Nyquist rate as illustrated in Section \ref{sec:setup}. 2- The CNN architecture outperforms the considered others, and hence, aside from single SNR selection using perfect estimates, we restrict our attention to this architecture. 3- The penalty incurred due to an inaccurate test SNR estimate used exclusively for training is not as significant as that of the modulation classification task.    
%Just like in the modulation classification dataset, single SNR selection in interference dataset leads to drastic decrease in training time while keeping accuracy high. For three considered deep neural network architectures: CNN, ResNet, and CLDNN, the results are similar to each other. We choose CNN to study in further sections because CNN outperforms all other architectures slightly. In the interference dataset, single SNR selection reduces the size of the training dataset to its $\frac{1}{21}$. Compared to uniform subsampling which reduces the size of the training dataset to its $\frac{1}{8}$, Figure \ref{fig_single:single_inter_multi} also shows that single SNR selection maintains higher accuracy while requiring less training data. From Figure \ref{fig_single:sensitivity_inter}, we can see that when test SNR becomes above $-2$ dB, the identification accuracy reaches around $100\%$ no matter if the SNR estimate was off by $4$ dB or $2$ dB. When the test SNR is below $-2$ dB, SNR estimate off by positive $2$ dB will give us better results compared to other estimates except for the correct SNR estimate. Overall, the results are pretty similar but still show a trend of positive offset being better than negative offset.
%\subsection{SNR Boosting}\label{sec:boosting}
\SetKwInput{KwInput}{Input}                % Set the Input
\SetKwInput{KwOutput}{Output}              % set the Output
\begin{algorithm}[H]
\DontPrintSemicolon
  \KwInput{Target SNR}
  \KwOutput{Selected Training SNR List}
  \While{Accuracy increase > 1 and remaining SNR list not empty}
    {
        \For{single SNR in remaining SNR list}
        {
            append single SNR training data to current set (without updating current set),
            train model, and evaluate on validation set
            
            \If{current accuracy > previous accuracy}
            {
                appending SNR $=$ single SNR
            }
        }
        \If{accuracy improvement}
        {add appending SNR to selected training SNR list
        
        remove appending SNR from remaining SNR list
        
        set Accuracy increase }
    }
\caption{SNR Boosting Algorithm}
\label{alg:Boosting}
\end{algorithm}

\begin{figure}[ht]
   \includegraphics[width=\linewidth,height=5cm]{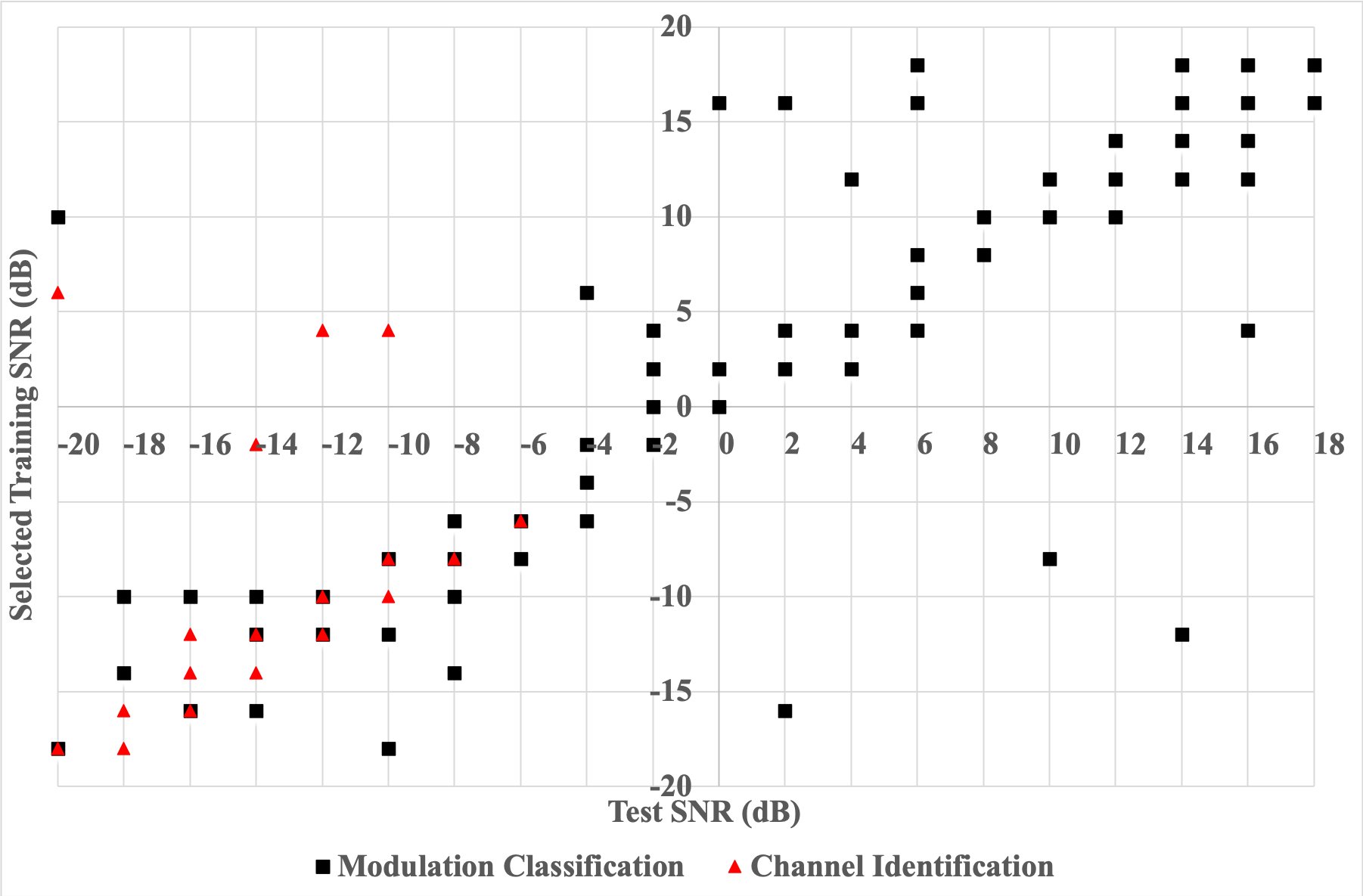}\par 
  \caption{SNR Set Selected by Boosting Algorithm.}
    \label{fig_single:greedy_set}
\end{figure}

\begin{figure}[ht]
\includegraphics[width=\linewidth, height=4cm]{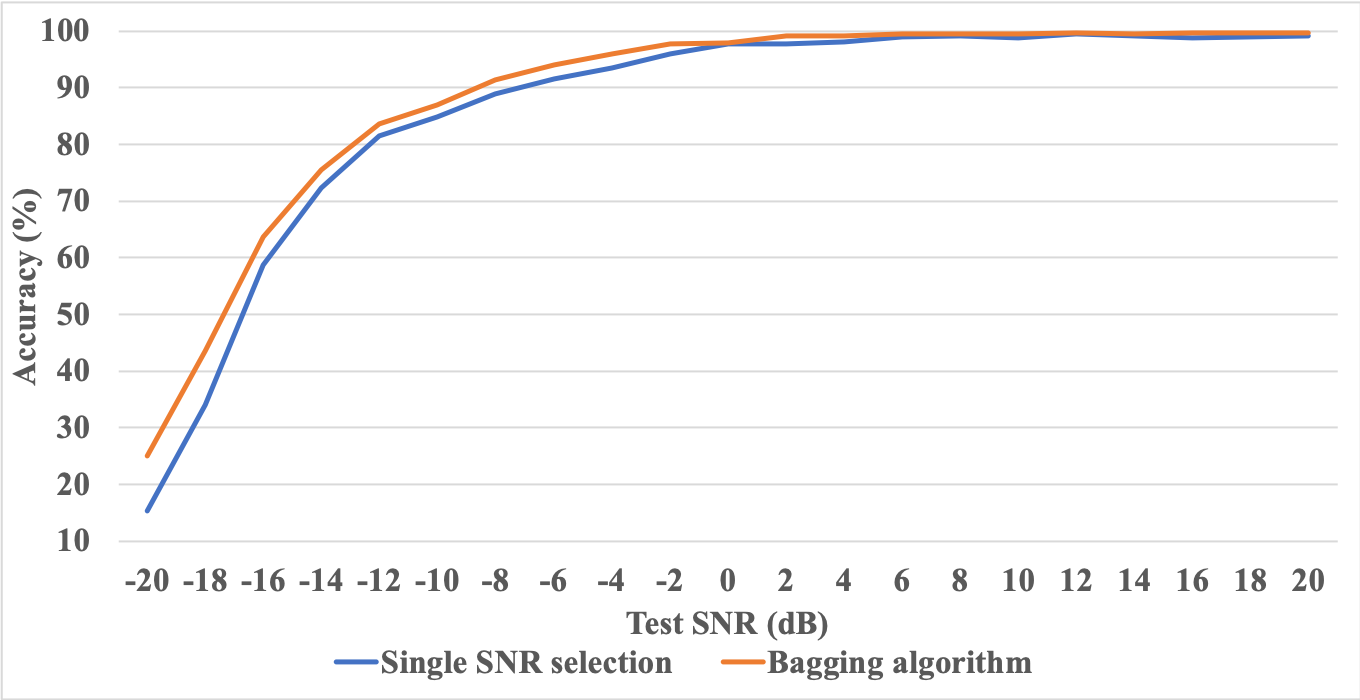}
\caption{Bagging using 5\% of one SNR training set size.}
\label{fig_greedy:bagging_5}
\end{figure}

\begin{figure}[ht]
\includegraphics[width=\linewidth, height=4cm]{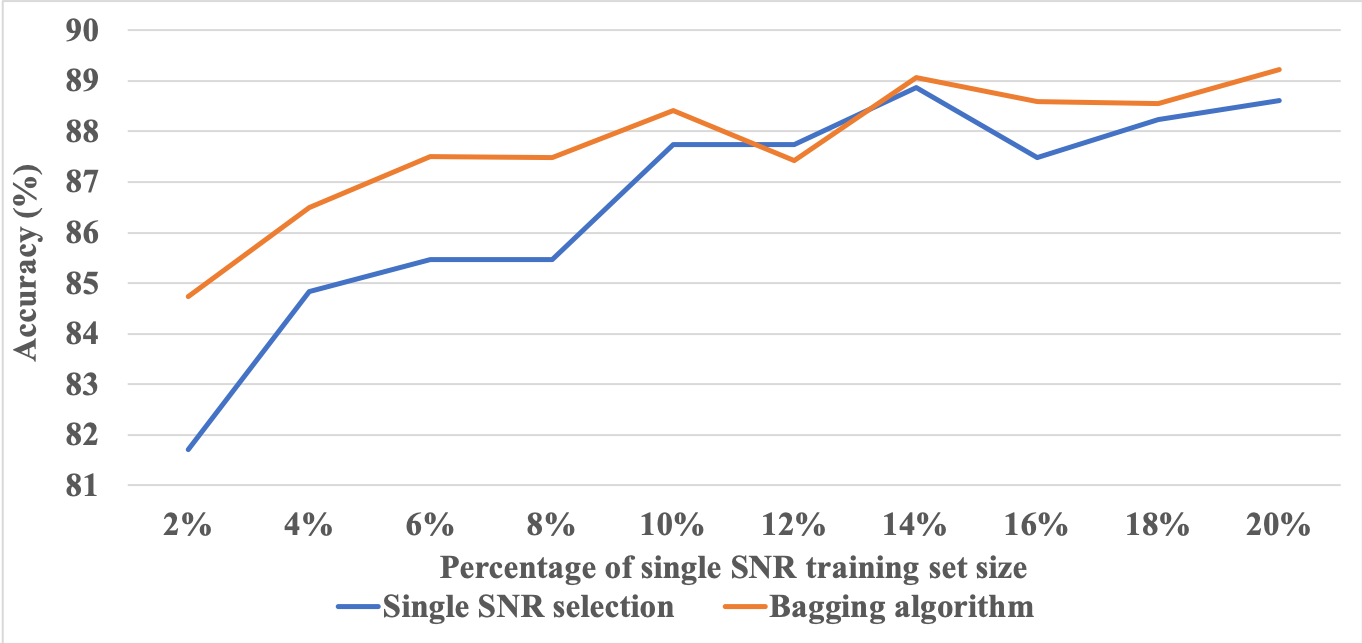}
\caption{Bagging at -$10$ dB  for Channel Identification.}
\label{fig_greedy:bagging_multi}
\end{figure}

\begin{figure}[ht]
\includegraphics[width=0.9\linewidth]{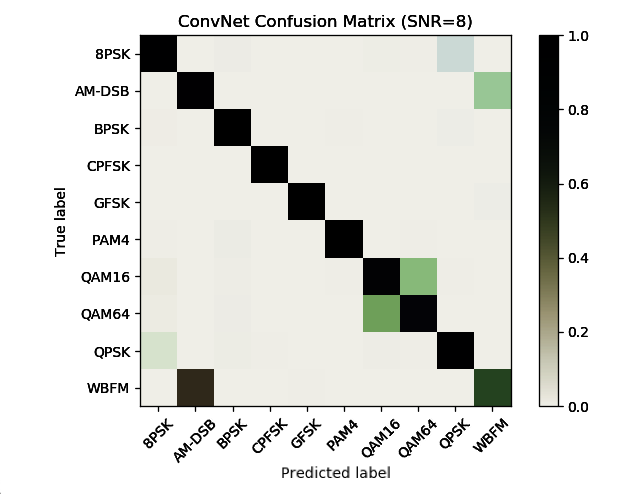}
\caption{Confusion matrices at 8 dB for modulation classification (ResNet). Different levels of red, green, and blue are used for representing the confusion matrix when training with: a) whole dataset, b) single SNR, and c) SNR Boosting, respectively. Note that the QAM16/QAM64 confusion is mostly present for a) and b) (green + red), while the QPSK/8PSK confusion is mostly present for b) and c) (green + blue).}
\label{fig_conf}
\end{figure}

\textbf{SNR Boosting:} The SNR Boosting Algorithm is shown in Algorithm \ref{alg:Boosting}. The validation set, consisting of 20\% of the set available for training, is used only for the selection of the training SNR values, and is later added back to the training set, when using these values to construct it. As observed in Figures \ref{fig_single:single_mod_multi} and \ref{fig_single:single_inter_multi}, training with the selected SNR set improves the model's performance. With the selected SNR set from our SNR boosting algorithm, the obtained accuracy is uniformly higher than - or very close to - that obtained when using the whole dataset. The training time is reduced significantly as the set typically consists of 3-5 SNR values, out of 20, for modulation classification, and 1-3 SNR values, out of 21, for low test SNR channel identification. The SNR values selected for training are marked in Figure \ref{fig_single:greedy_set}. For low SNR values, it is rarely the case that training with significantly higher SNR values is beneficial; this may explain the good performance of SNR selection at low SNR, since in this regime, all the training data needed for good performance has high noise levels, and hence, only a small subset - corresponding to the test SNR value - suffices to achieve that performance. We further observe from the confusion matrices in Figure \ref{fig_conf} how at the high test SNR value of 8 dB, one extra training SNR value (10 dB) can lead to significant performance improvement by resolving the confusion between the QAM16/QAM64 classes. We finally observe from Figure \ref{fig_single:single_mod_multi} and Figure \ref{fig_conf} that although the accuracy obtained through training with the whole dataset and the boosting set is similar, the errors emanate from confusions between different class pairs, suggesting the potential for better performance through other - potentially non-greedy - boosting algorithms, which we will investigate in future work.

%\begin{figure*}[ht]
%\begin{multicols}{3}

%    \hfill
%    \subfloat[Trained with whole dataset]{
%    \includegraphics[width=1.15\linewidth]{conf_all.png}
%    \label{fig_conf:all}
%    }

%    \hfill
%    \subfloat[Trained with single SNR selection]{
%    \includegraphics[width=1.15\linewidth]{conf_single.png}
%    \label{fig_conf:single}
%    }

%    \hfill
%    \subfloat[Trained with SNR boosting]{
%    \includegraphics[width=1.15\linewidth]{conf_greedy.png}
%    \label{fig_conf:boosting}
%    }
%\end{multicols}
%\caption{Confusion matrix at 8 dB for modulation classification (ResNet).}
%\label{fig_conf}
%\end{figure*}

\section{Discussion}
%\subsection{SNR Bagging}

%\begin{figure}[ht]
%\hfill
%\subfloat[Bagging using 5\% of one SNR training set size for Channel Identification.]{
%\includegraphics[width=\linewidth, height=4cm]{bagging_inter_5.png}
%\label{fig_greedy:bagging_5}
%}

%\hfill
%\subfloat[Bagging at -$10$ dB  for Channel Identification.]{
%\includegraphics[width=\linewidth, height=3.5cm]{bagging_inter_multi.png}
%\label{fig_greedy:bagging_multi}
%}

%\caption{Bagging at -$10$ dB  for Channel Identification.}
%\label{fig_greedy}
%\end{figure}

\textbf{SNR Bagging:} To improve generalization performance, we implemented a bootstrap aggregating (Bagging) algorithm that relies on training three identical models using independently and uniformly sampled training sets from the sets corresponding to the test SNR value as well as the two adjacent values. During inference, we hold a vote among the three models. We compare the test accuracy to that obtained by single SNR selection using a training set of the same size. We observe from Figures~\ref{fig_greedy:bagging_5} and \ref{fig_greedy:bagging_multi} that noticeable improvements in test accuracy are obtained for channel identification for smaller training sets and lower SNR values. We believe that generalization becomes a dominant factor in determining the learning performance in presence of a scarcity of data that carries clear patterns. 

%\subsection{Larger Datasets}
\textbf{Larger Datasets:} We performed the single SNR selection and sensitivity studies using the larger RadioML2018.01A dataset, that was first used in \cite{newdataset}, and has 24 modulation types corresponding to both synthetic and over-the-air received samples, and 1024 samples per input vector. The ResNet of Table \ref{table: neural network architecture for modulation} was used with slight modifications to accommodate the new dimensions. Similar insights were obtained, which demonstrate the generality of our conclusions. Specifically, we found the conclusions related to the superior relative performance at low SNR for SNR selection, as well as the superior performance of positive erroneous estimates to hold.

\ifCLASSOPTIONcaptionsoff
  \newpage
\fi

\bibliographystyle{IEEEtran} 
\bibliography{ElGamal_WCL2020-0502}

\end{document}